\pdfoutput=1
\documentclass[11pt]{article}
\usepackage[]{acl}
\usepackage{times}
\usepackage{latexsym}
\usepackage{url}
\usepackage{multirow}
\usepackage{graphicx}
\usepackage{amssymb}
\usepackage{pifont}
\newcommand{\cmark}{\ding{51}}%
\newcommand{\xmark}{\ding{55}}%
\usepackage{colortbl}
\definecolor{dGray}{gray}{.6}
\definecolor{mGray}{gray}{.82}
\definecolor{lGray}{gray}{.92}
\usepackage[T1]{fontenc}
\usepackage[utf8]{inputenc}
\usepackage{microtype}
\usepackage{inconsolata}
\usepackage{subcaption}

\title{Enhancing Data Quality through Simple De-duplication: Navigating Responsible Computational Social Science Research}

\author{Yida Mu \qquad Mali Jin \qquad Xingyi Song \qquad Nikolaos Aletras\\
  School of Computer Science, The University of Sheffield \\
  \texttt{\{y.mu, m.jin, x.song, n.aletras\}@sheffield.ac.uk} \\}

\begin{document}
\maketitle

\begin{abstract}

Research in natural language processing (NLP) for Computational Social Science (CSS) heavily relies on data from social media platforms. This data plays a crucial role in the development of models for analysing socio-linguistic phenomena within online communities. 
In this work, we conduct an in-depth examination of 20 datasets extensively used in NLP for CSS to comprehensively examine data quality. 
Our analysis reveals that social media datasets exhibit varying levels of data duplication. Consequently, this gives rise to challenges like label inconsistencies and data leakage, compromising the reliability of models. Our findings also suggest that data duplication has an impact on the current claims of state-of-the-art performance, potentially leading to an overestimation of model effectiveness in real-world scenarios.
Finally, we propose new protocols and best practices for improving dataset development from social media data and its usage.

\end{abstract}

\section{Introduction}

Research in natural language processing (NLP) for Computational Social Science (CSS) aims to analyze social behavior and sociolinguistic phenomena with computational methods on a large scale \citep{plank2015personality,guntuku2019twitter}. This relies upon using vast amounts of user-generated content in social media such as Twitter/X and Weibo \citep{edelmann2020computational}.

Social media data exhibits distinctive characteristics such as rapid and continual topic evolution \citep{saha2012learning,yuan2013and}. Prior research has focused on introducing and developing novel resources including new tasks and datasets~\cite{founta2018large,lazer2020computational, hofman2021integrating}.  
Social media posts often contain a significant amount of near-duplicate or even identical content. For example, during social emergencies (e.g., COVID-19 pandemic), users tend to repeatedly post about the same topics (e.g., COVID-19 vaccination), resulting in a proliferation of near-duplicate content over a very short period of time \citep{ferrara2020types,zhang2022social}. Furthermore, the presence of social bots (e.g., automatically generated posts from third-party applications) amplifies the rapid generation of near-duplicate content on these platforms, e.g., in political campaigns \citep{bessi2016social,stella2018bots}. 

Data quality is paramount for reliable and effective model performance. It includes various dimensions such as accuracy, completeness and consistency \citep{vidgen2020directions,budach2022effects}. Ensuring high data quality usually involves careful data management and processing such as data validation and cleaning \citep{denny2018text,breck2019data,groger2021there,li2021cleanml}. Previous works have examined and deduplication of samples in image datasets \citep{alam2017image4act}, generic real-word datasets (e.g., movie, restaurant) \cite{li2021cleanml} and language modeling datasets (e.g., Wikipedia) \citep{lee2022deduplicating}. However, the quality of existing CSS datasets in NLP has been under-explored. 

The goal of this paper is to conduct a large scale meta-analysis of current NLP for CSS datasets by considering the potential noise caused by duplicated text samples. To this end, we perform an in-depth re-examination of 20 social media datasets across various tasks such as offensive language detection and misinformation detection.\footnote{Data and code: \url{https://github.com/YIDAMU/Clean_CSS}} Our main contributions are as follows:

\begin{table*}[!t]
\resizebox{\textwidth}{!}{
\begin{tabular}{|l|l|c|c|}
\hline
\rowcolor{dGray}
\multicolumn{1}{|c|}{\cellcolor{dGray}\textbf{Tweet\_ID}} &
  \multicolumn{1}{c|}{\cellcolor{dGray}\textbf{Samples}} &
  \textbf{Label} &
  \multicolumn{1}{l|}{\cellcolor{dGray}\textbf{Consequence}} \\ \hline
\rowcolor[HTML]{FFFFFF} 
129839* &
  @USER @USER donald trump's lessons for republicans: consequences for lying... &
  Neutral &
  \cellcolor[HTML]{FFFFFF} \\ \cline{1-3}
\rowcolor[HTML]{FFFFFF} 
129840* &
  @USER @USER donald trump's lessons for republicans: consequences for lying... &
  Against &
  \multirow{-2}{*}{\cellcolor[HTML]{FFFFFF}\textbf{\begin{tabular}[c]{@{}c@{}}Inconsistent \\ Labeling\end{tabular}}} \\ \hline
\rowcolor[HTML]{FFFFFF} 
132605* &
  \begin{tabular}[c]{@{}l@{}}@USER FIGHT AGAINST TYRANNY!!! The coronavirus vaccine is a NEW \\ technology RNA vaccine.  It LITERALLY changes your DNA!!!! {[}Emojis{]}\end{tabular} &
  \begin{tabular}[c]{@{}c@{}}Anti. \\ Vaxx\end{tabular} &
  \cellcolor[HTML]{FFFFFF} \\ \cline{1-3}
\rowcolor[HTML]{FFFFFF} 
132607* &
  \begin{tabular}[c]{@{}l@{}}@USER FIGHT AGAINST TYRANNY!!! The coronavirus vaccine is a NEW \\ technology RNA vaccine.  It LITERALLY changes your DNA!!!! {[}Emojis{]}\end{tabular} &
  \begin{tabular}[c]{@{}c@{}}Anti. \\ Vaxx\end{tabular} &
  \multirow{-2}{*}{\cellcolor[HTML]{FFFFFF}\textbf{\begin{tabular}[c]{@{}c@{}}Label \\ Leakage\end{tabular}}} \\ \hline
\rowcolor[HTML]{FFFFFF} 
\begin{tabular}[c]{@{}l@{}}131120*\\ (Dupli.) $\dagger$ \end{tabular} &
  @USER trump &
  Neutral &
  \textbf{\begin{tabular}[c]{@{}c@{}}Label\\ Leakage\end{tabular}} \\ \hline
\rowcolor[HTML]{FFFFFF} 
407156* &
  \begin{tabular}[c]{@{}l@{}}r.i.p to the driver \textcolor{red}{who} died with paul walker that no one cares about because \\ he wasn’t famous \textcolor{red}{omg:(}\end{tabular} &
  \begin{tabular}[c]{@{}c@{}}True\\ Rumor\end{tabular} &
  \cellcolor[HTML]{FFFFFF} \\ \cline{1-3}
\rowcolor[HTML]{FFFFFF} 
407164* &
  \begin{tabular}[c]{@{}l@{}}r.i.p to the driver \textcolor{red}{that} died with paul walker that no one cares about because \\ he wasn't famous.\end{tabular} &
  \cellcolor[HTML]{FFFFFF}\begin{tabular}[c]{@{}c@{}}True\\ Rumor\end{tabular} &
  \multirow{-2}{*}{\cellcolor[HTML]{FFFFFF}\textbf{\begin{tabular}[c]{@{}c@{}}Label\\ Leakage\end{tabular}}} \\ \hline
\end{tabular}
}
\caption{ Examples of duplicate or near-duplicate samples in real-world social datasets \citep{ma2017detect,cotfas2021longest,kawintiranon2021knowledge} as well as their potential consequence. $\dagger$ denotes duplicate tweet ids in the dataset.} 
\label{tab:duplicates_examples}
\end{table*}

\begin{itemize}

    \item  Our systematic analysis shows that most of the examined social media datasets contain noise (e.g., duplicate and near-duplicate samples) despite the data cleaning process claimed by the developers (see examples in Table \ref{tab:duplicates_examples}). 

    \item We explore the impact of data duplication on model performance in various CSS tasks. We observe an overestimation of model performance in cases where duplicate or near-duplicate samples remain unfiltered. We find that the presence of duplicate samples results in label inconsistencies and data leakage, potentially causing unreliable model predictions.

    \item We propose a new data pre-processing protocol for a more responsible and effective use of social media resources and advocate for a `minor revision' of the existing author checklist from major CSS communities.

\end{itemize}
Note that this study does not aim to criticize dataset creators. Instead, it acknowledges the crucial role these datasets play in advancing the field of CSS and seeks to provide constructive recommendations for improving responsible research and practices.

\section{Related Work}

\subsection{Data Quality and Model Performance}

Data quality is paramount in NLP tasks as the performance and reliability of models heavily depend on the quality of the training data. Several factors contribute to data quality such as accuracy, persistence, completeness, consistency and relevance \citep{zubiaga2018longitudinal,assenmacher2020end,koch2021reduced,budach2022effects}. \citet{budach2022effects} investigated the impact of different data quality dimensions on the performance of various machine learning algorithms covering classification, regression and clustering tasks. Similarly,  \citet{barry2023impact} explored the relationship between the quality of the training data and the fairness of model outputs on image classification across a range of algorithms. Furthermore, \citet{li2021cleanml} examined the impact of improving data quality on classification algorithm performance, resulting in the CleanML benchmark. Their work highlighted the importance of data cleaning methods such as addressing missing values and correcting mislabels in enhancing classifier predictions.

\subsection{Data Preprocessing}

Ensuring data quality in NLP often involves rigorous data preprocessing. Dealing with social media datasets often comes with unique challenges due to their unstructured nature and noisy text \citep{symeonidis2018comparative}.

The majority of CSS research follows a standard pre-processing pipeline for social media content \citep{nguyen2020bertweet,antypas2023supertweeteval}. Social media text often contains user mentions, URLs, hashtags, emojis, and other non-standard symbols. Pre-processing steps typically involve removing or replacing these tokens with generic placeholders. Additionally, social media language is informal, featuring slang, abbreviations, and misspellings, necessitating text normalization techniques such as lowercase conversion and spelling correction \citep{baldwin2015depth,naseem2021survey}.

Previous studies have investigated the impact of various pre-processing strategies on model performance in different CSS downstream tasks such as sentiment analysis \citep{krouska2016effect,jianqiang2017comparison,mahilraj2020text} and opinion mining \citep{dos2014role,gull2016pre}. For example, \citet{symeonidis2018comparative} examined up to 16 different pre-processing strategies for Twitter data and found that techniques such as lemmatization and removing numbers in the text can enhance the predictive performance of transformer-based models in sentiment analysis.

\subsection{Impact of Data Duplication}
Existing language modeling datasets contain many duplicate and near-duplicate samples due to overlapping sources in the training corpus, which has emerged as a significant concern in recent research. Studies have demonstrated that large language models are vulnerable to privacy attacks due to the presence of duplicate sequences in commonly used training datasets \citep{kandpal2022deduplicating}. These sequences, when present multiple times in the training data, are regenerated at much higher frequencies by trained models, posing privacy risks. Furthermore, data duplication in language modeling leads to verbatim text duplication in model outputs \citep{lee2022deduplicating}. Efforts to mitigate this issue have focused on deduplicating training datasets, resulting in models emitting memorized text less frequently and requiring fewer training steps for comparable or improved performance \citep{lee2022deduplicating}.

In this study, we focus on examining and mitigating the potential issue of noise data (i.e., duplicate data) within existing CSS datasets, which has often been overlooked. To our knowledge, a thorough re-evaluation of of the quality of social media datasets has not been extensively conducted.

\section{Tasks \& Datasets}
We evaluate a comprehensive collection of datasets that adequately cover a broad range of prevalent tasks in CSS. Based on recent publications in NLP and CSS venues (e.g., *ACL and ICWSM), we choose datasets from four main CSS tasks following previous work \citep{barbieri2020tweeteval,ziems2023can,antypas2023supertweeteval,mu2024navigating}: (i) Offensive Language Detection, (ii) Misinformation Detection, (iii) Speech Act Detection \& Sentiment Analysis, and (iv) Stance Detection.

\subsection{Offensive Language Detection}
Offensive language detection refers to the process of automatically identifying content that contains hate speech, harassment, profanity, or other forms of inappropriate language towards individuals, groups, or events \citep{chen2012detecting,davidson2017automated}. It is particularly relevant for content moderation in online platforms and social media \citep{nobata2016abusive}.

For this task, we opt for the following datasets:
\texttt{WASEEM} \citep{waseem2016hateful}, \texttt{TBO} \citep{zampieri2023target}, \texttt{OLID} \citep{davidson2017automated}, \texttt{FOUNTA} \citep{founta2018large} and \texttt{HateEval'19} \citep{garibo-i-orts-2019-multilingual}.

\subsection{Misinformation Detection}
Misinformation detection in social media involves the use of algorithms to analyze and identify false or misleading information (e.g., fake news and false rumors) generated or diffused by end users
\citep{shu2017fake,zubiaga2018detection}. These approaches typically rely on linguistic patterns (e.g., hand-craft features), source credibility (e.g., unreliable news sources), and contextual information (e.g., propagation network) to differentiate between true information and misinformation \citep{rashkin2017truth,tian2022duck}.

We conduct evaluation on five popular datasets covering different languages and social media platforms: \texttt{Twitter 15}, \texttt{Twitter 16} \citep{ma2017detect}, \texttt{PHEME} \citep{zubiaga2016analysing}, \texttt{Weibo 16} \citep{ma2016detecting}, \texttt{Weibo 20} \citep{rao-etal-2021-stanker}. 

\subsection{Speech Act Detection \& Sentiment Analysis}
Speech act detection and sentiment analysis deal with the detection and analysis of actions (e.g., requesting and complaining) and affecting content within texts. In recent years, there has been growing attention to automatically identifying speech acts and sentiment analysis in social media~\citep{preotiuc-pietro-etal-2019-automatically,farha2022semeval,zhang-etal-2022-curriculum}. These two tasks are closely related as detecting one usually involves the other~\citep{saha-etal-2021-towards}.

We choose the following speech acts and sentiment analysis datasets: \texttt{Complaint}~\cite{preotiuc-pietro-etal-2019-automatically}, \texttt{SemEval-2022 Task 6 Sarcasm} \citep{farha2022semeval}, \texttt{Bragging}~\cite{jin-etal-2022-automatic}, \texttt{Parody} \citep{maronikolakis2020analyzing} and \texttt{SemEval-2017 Task 4 Sentiment} \citep{rosenthal2019semeval}.

\begin{table*}[!t]
\centering
\resizebox{0.95\textwidth}{!}{%
\begin{tabular}{|lccccrrr|}
\hline
\rowcolor{dGray} 
\multicolumn{1}{|c|}{\cellcolor{dGray}\textbf{\begin{tabular}[c]{@{}c@{}}Domain /\\ Dataset\end{tabular}}} &
  \multicolumn{1}{c|}{\cellcolor{dGray}\textbf{\begin{tabular}[c]{@{}c@{}}Source /\\  Language\end{tabular}}} &
  \multicolumn{1}{c|}{\cellcolor{dGray}\textbf{\begin{tabular}[c]{@{}c@{}}Mean\\ Tokens\end{tabular}}} &
  \multicolumn{1}{c|}{\cellcolor{dGray}\textbf{\begin{tabular}[c]{@{}c@{}}Self-claimed\\ Deduplication\end{tabular}}} &
  \multicolumn{1}{c|}{\cellcolor{dGray}\textbf{\# of Post}} &
  \multicolumn{1}{c|}{\cellcolor{dGray}\textbf{\begin{tabular}[c]{@{}c@{}}\# of Distinct\\ / Ratio\%\end{tabular}}} &
  \multicolumn{1}{c|}{\textbf{\begin{tabular}[c]{@{}c@{}}\# of Distinct\\ after Pre-proc.\\ / Ratio \%\end{tabular}}} &
  \textbf{\begin{tabular}[c]{@{}c@{}}\ \# of Distinct \\ after Removing \\  Near-Dupli. / Ratio\%\end{tabular}} \\ \hline
\rowcolor[HTML]{FFCE93} 
\multicolumn{8}{|c|}{\cellcolor{mGray}\textbf{Offensive Language Analysis}} \\ \hline
\multicolumn{1}{|l|}{\texttt{\textbf{WASEEM}}} &
  \multicolumn{1}{c|}{Twitter/En} &
  \multicolumn{1}{c|}{15} &
  \multicolumn{1}{c|}{\xmark} &
  \multicolumn{1}{c|}{16,909} &
  \multicolumn{1}{r|}{16,851 / 99.7\%} &
  \multicolumn{1}{r|}{16,568 / 98.0\%} &
  13,364 / 79.0\% \\ \hline
\multicolumn{1}{|l|}{\texttt{\textbf{TBO}}} &
  \multicolumn{1}{c|}{Twitter/En} &
  \multicolumn{1}{c|}{17} &
  \multicolumn{1}{c|}{\cmark} &
  \multicolumn{1}{c|}{\cellcolor{lGray}4,000} &
  \multicolumn{1}{r|}{\cellcolor{lGray}3,998 / 99.9\%} &
  \multicolumn{1}{r|}{\cellcolor{lGray}3,998 / 99.9\%} &
  3,946 / 98.7\% \\ \hline
\multicolumn{1}{|l|}{\texttt{\textbf{OLID}}} &
  \multicolumn{1}{c|}{Twitter/En} &
  \multicolumn{1}{c|}{22} &
  \multicolumn{1}{c|}{\xmark} &
  \multicolumn{1}{c|}{14,100} &
  \multicolumn{1}{r|}{14,052 / 99.7\%} &
  \multicolumn{1}{r|}{14,031 / 99.5\%} &
  1,1509 / 81.6\% \\ \hline
\multicolumn{1}{|l|}{\texttt{\textbf{FOUNTA}}} &
  \multicolumn{1}{c|}{Twitter/En} &
  \multicolumn{1}{c|}{17} &
  \multicolumn{1}{c|}{\xmark} &
  \multicolumn{1}{c|}{99,996} &
  \multicolumn{1}{r|}{91,940 / 91.9\%} &
  \multicolumn{1}{r|}{87,291 / 87.3\%} &
  88,263 / 81.5\% \\ \hline
\multicolumn{1}{|l|}{\texttt{\textbf{HatEval'19}}} &
  \multicolumn{1}{c|}{Twitter/En} &
  \multicolumn{1}{c|}{22} &
  \multicolumn{1}{c|}{\xmark} &
  \multicolumn{1}{c|}{19,600} &
  \multicolumn{1}{r|}{19,342 / 98.7\%} &
  \multicolumn{1}{r|}{19,267 / 98.3\%} &
  17,851 / 91.0\% \\ \hline
\rowcolor[HTML]{FFCE93} 
\multicolumn{8}{|c|}{\cellcolor{mGray}\textbf{Misinformation Detection}} \\ \hline
\multicolumn{1}{|l|}{\texttt{\textbf{Twitter 15}}} &
  \multicolumn{1}{c|}{Twitter/En} &
  \multicolumn{1}{c|}{15} &
  \multicolumn{1}{c|}{\xmark} &
  \multicolumn{1}{c|}{1,490} &
  \multicolumn{1}{r|}{1,428 / 95.8\%} &
  \multicolumn{1}{r|}{1,428 / 95.8\%} &
  1,345 / 90.2\% \\ \hline
\multicolumn{1}{|l|}{\texttt{\textbf{Twitter 16}}} &
  \multicolumn{1}{c|}{Twitter/En} &
  \multicolumn{1}{c|}{15} &
  \multicolumn{1}{c|}{\xmark} &
  \multicolumn{1}{c|}{818} &
  \multicolumn{1}{r|}{761 / 93.0\%} &
  \multicolumn{1}{r|}{761 / 93.0\%} &
  740 / 90.5\% \\ \hline
\multicolumn{1}{|l|}{\texttt{\textbf{PHEME}}} &
  \multicolumn{1}{c|}{Twitter/En} &
  \multicolumn{1}{c|}{16} &
  \multicolumn{1}{c|}{\xmark} &
  \multicolumn{1}{c|}{5,802} &
  \multicolumn{1}{r|}{5,789 / 99.8\%} &
  \multicolumn{1}{r|}{5,694 / 98.1\%} &
  5,236 / 90.2\% \\ \hline
\multicolumn{1}{|l|}{\texttt{\textbf{Weibo}}} &
  \multicolumn{1}{c|}{Weibo/Zh} &
  \multicolumn{1}{c|}{99} &
  \multicolumn{1}{c|}{\xmark} &
  \multicolumn{1}{c|}{4,664} &
  \multicolumn{1}{r|}{4,516 / 96.8\%} &
  \multicolumn{1}{r|}{4,501 / 96.5\%} &
  3,322 / 71.2\% \\ \hline
\multicolumn{1}{|l|}{\texttt{\textbf{STANKER}}} &
  \multicolumn{1}{c|}{Weibo/Zh} &
  \multicolumn{1}{c|}{71} &
  \multicolumn{1}{c|}{\xmark} &
  \multicolumn{1}{c|}{\cellcolor{lGray}6,068} &
  \multicolumn{1}{r|}{\cellcolor{lGray}6,040 / 99.5\%} &
  \multicolumn{1}{r|}{6,006 / 99.0\%} &
  4,703 77.5\% \\ \hline
\rowcolor[HTML]{FFCE93} 
\multicolumn{8}{|c|}{\cellcolor{mGray}\textbf{Speech Act \& Sentiment Analysis}} \\ \hline
\multicolumn{1}{|l|}{\texttt{\textbf{Complaint}}} &
  \multicolumn{1}{c|}{Twitter/En} &
  \multicolumn{1}{c|}{15} &
  \multicolumn{1}{c|}{\cmark} &
  \multicolumn{1}{c|}{\cellcolor{lGray}3,449} &
  \multicolumn{1}{r|}{\cellcolor{lGray}3,449 / 100.0\%} &
  \multicolumn{1}{r|}{3,408 / 98.8\%} &
  2,846 / 82.5\% \\ \hline
\multicolumn{1}{|l|}{\texttt{\textbf{Sarcasm}}} &
  \multicolumn{1}{c|}{Twitter/En} &
  \multicolumn{1}{c|}{18} &
  \multicolumn{1}{c|}{\xmark} &
  \multicolumn{1}{c|}{4,868} &
  \multicolumn{1}{r|}{4,851 / 99.7\%} &
  \multicolumn{1}{r|}{4,849 / 99.6\%} &
  4,442 / 91.2\% \\ \hline
\multicolumn{1}{|l|}{\texttt{\textbf{Bragging}}} &
  \multicolumn{1}{c|}{Twitter/En} &
  \multicolumn{1}{c|}{22} &
  \multicolumn{1}{c|}{\xmark} &
  \multicolumn{1}{c|}{6,696} &
  \multicolumn{1}{r|}{6,643 / 99.2\%} &
  \multicolumn{1}{r|}{6,636 / 99.1\%} &
  5,979 / 89.2\% \\ \hline
\multicolumn{1}{|l|}{\texttt{\textbf{Parody}}} &
  \multicolumn{1}{c|}{Twitter/En} &
  \multicolumn{1}{c|}{29} &
  \multicolumn{1}{c|}{\xmark} &
  \multicolumn{1}{c|}{\cellcolor{lGray}46,622} &
  \multicolumn{1}{r|}{\cellcolor{lGray}46,587 / 99.9\%} &
  \multicolumn{1}{r|}{46,024 / 98.7\%} &
  43,591 / 93.5\% \\ \hline
\multicolumn{1}{|l|}{\texttt{\textbf{Sentiment}}} &
  \multicolumn{1}{c|}{Twitter/En} &
  \multicolumn{1}{c|}{18} &
  \multicolumn{1}{c|}{\xmark} &
  \multicolumn{1}{c|}{\cellcolor{lGray}59,899} &
  \multicolumn{1}{r|}{\cellcolor{lGray}59,870 / 99.9\%} &
  \multicolumn{1}{r|}{\cellcolor{lGray}59,836 / 99.9\%} &
  58,536 / 97.7\% \\ \hline
\rowcolor[HTML]{FFCE93} 
\multicolumn{8}{|c|}{\cellcolor{mGray}\textbf{Stance Detection}} \\ \hline
\multicolumn{1}{|l|}{\texttt{\textbf{CovidVaxx}}} &
  \multicolumn{1}{c|}{Twitter/En} &
  \multicolumn{1}{c|}{25} &
  \multicolumn{1}{c|}{\xmark} &
  \multicolumn{1}{c|}{\cellcolor{lGray}2,792} &
  \multicolumn{1}{r|}{\cellcolor{lGray}2,787 / 99.8\%} &
  \multicolumn{1}{r|}{2,740 / 98.1\%} &
  2,567 / 91.9\% \\ \hline
\multicolumn{1}{|l|}{\texttt{\textbf{RumorEval}}} &
  \multicolumn{1}{c|}{Twitter/En} &
  \multicolumn{1}{c|}{33} &
  \multicolumn{1}{c|}{\xmark} &
  \multicolumn{1}{c|}{5,568} &
  \multicolumn{1}{r|}{5,467 / 98.2\%} &
  \multicolumn{1}{r|}{5,467 / 98.2\%} &
  4,284 / 76.9\% \\ \hline
\multicolumn{1}{|l|}{\texttt{\textbf{US-Election}}} &
  \multicolumn{1}{c|}{Twitter/En} &
  \multicolumn{1}{c|}{25} &
  \multicolumn{1}{c|}{\cmark} &
  \multicolumn{1}{c|}{\cellcolor{lGray}2,500} &
  \multicolumn{1}{r|}{\cellcolor{lGray}2,498 / 99.9\%} &
  \multicolumn{1}{r|}{\cellcolor{lGray}2,498 / 99.9\%} &
  2,397 / 95.9\% \\ \hline
\multicolumn{1}{|l|}{\texttt{\textbf{P-Stance}}} &
  \multicolumn{1}{c|}{Twitter/En} &
  \multicolumn{1}{c|}{30} &
  \multicolumn{1}{c|}{\cmark} &
  \multicolumn{1}{c|}{\cellcolor{lGray}21,574} &
  \multicolumn{1}{r|}{\cellcolor{lGray}21,571 / 99.9\%} &
  \multicolumn{1}{r|}{\cellcolor{lGray}21,571 / 99.9\%} &
  21,551 / 99.8\% \\ \hline
\multicolumn{1}{|l|}{\texttt{\textbf{SemEval’16}}} &
  \multicolumn{1}{c|}{Twitter/En} &
  \multicolumn{1}{c|}{17} &
  \multicolumn{1}{c|}{\cmark} &
  \multicolumn{1}{c|}{\cellcolor{lGray}4,063} &
  \multicolumn{1}{r|}{\cellcolor{lGray}4,063 / 100.0\%} &
  \multicolumn{1}{r|}{4,048 / 99.6\%} &
  3,926 / 96.6\% \\ \hline
\end{tabular}%
}
\caption{Dataset Specifications. Cells in light grey indicate no or minor reduction (i.e., less than 0.1\%) from duplicate samples. Note that some social media datasets have been pre-processed (e.g., by replacing @USER and URL with special tokens) before being publicly available such as Twitter 15 and Twitter 16. This results in the same values for \textit{\# of Distinct} and \textit{\# of Distinct Posts after Pre-proc.}.}
\label{tab:datasets}
\end{table*}

\subsection{Stance Detection}
Stance detection involves using computational approaches to automatically identify and classify a person's or a group's perspective or attitude towards a specific target, such as events (e.g., COVID-19 vaccination) or individuals (e.g., politicians) \citep{kuccuk2020stance,mu2023vaxxhesitancy}. 

We choose five datasets for the stance detection task including \texttt{COVID-19 Vaccine Stance} \citep{cotfas2021longest}, \texttt{SemEval-2019 Task 7 Rumor Stance} \citep{gorrell-etal-2019-semeval}, \texttt{US-Election}~\citep{kawintiranon2021knowledge}, \texttt{P-Stance}~\citep{li2021p}, and \texttt{Semeval-2016 Task 6}~\citep{mohammad2016semeval}.

\subsection{Criteria for Dataset Selection}
For each domain, we select five representative datasets based on criteria that include: (i) popularity, as indicated by citation counts; (ii) shared tasks from SemEval\footnote{International Workshop on Semantic Evaluation: \url{https://semeval.github.io/}} and (iii) newly developed datasets (see Table \ref{tab:datasets} for tasks and related datasets). 

For example, \texttt{WASEEM} \citep{waseem2016hateful} is one of the most popular and earliest dataset for studying online hate speech. Similarly, \texttt{Twitter 15 \& 16} \citep{ma2017detect} and \texttt{PHEME} \citep{zubiaga2016analysing} have been widely used in recent work on computational rumor detection, as shown in \cite{mu2023s}.
Moreover, we consider SemEval datasets as they are generally of high interest to the NLP/CSS community. These shared tasks often release leaderboards for ranking participants, underscoring the importance of using a clean dataset. 
We also use some newly developed datasets that cover specific linguistic phenomena, such as \texttt{Bragging} \citep{jin-etal-2022-automatic}, which is related to computational pragmatics.

\section{Examining Dataset Quality}
\label{statistics}
We focus on examining duplicate or near-duplicate samples in datasets to assess data quality. 

\subsection{Dataset Specifications}
To assess the duplication and near-duplication quantity, we report the following specifications:

\begin{itemize}
    \item  \textbf{Self-claimed deduplication.} We report this feature by manually reviewing the source paper of the dataset. The purpose of this is to understand whether the original authors have performed data preprocessing and how data has been preprocessed. The label `\cmark' indicates that the developers have clearly mentioned their implementation of deduplicating in their datasets. 
    
    \item \textbf{Number of posts in the datasets.} We present the number of samples in the original datasets which were obtained either from authors or through the links provided in source papers.\footnote{The size of some datasets varies from the original one due to recollection.}
    
    \item \textbf{Number of distinct posts.} We display the number of \textit{unique posts} in the original datasets. In this step, we only filter out samples that contain the same content.

    \item \textbf{Number of distinct posts after removing duplicates.} We present the number of distinct posts after preprocessing, i.e., replacing @USER and URL tokens with unified tokens. 
    
    \item  \textbf{Number of distinct posts after removing near-duplicates.} We further employ the Levenshtein distance \citep{levenshtein1966binary} to calculate the similarity between each sample and filter out near-duplicate samples after replacing @USER and URL tokens (threshold = 20). Note that this set of data excludes both duplicate and near-duplicate samples.
    
\end{itemize}

\subsection{Results and Discussion}
Table \ref{tab:datasets} presents the above features as well as basic information including source platforms, language and number of average tokens (word and character level for English and Chinese respectively). Generally, we observe that most existing social media datasets (18 out of 20 in total) contain duplicate samples. Replacing special tokens reveals additional duplicate samples in the majority of the datasets (17 out of 20 in total). Furthermore, we notice that datasets with high duplicate rates are usually developed through a keyword-based sampling method.

Additionally, we note that only a small fraction of papers claim that they have performed a deduplication process, e.g., \texttt{TBO} \citep{zampieri2023target} and Complaint \citep{preotiuc-pietro-etal-2019-automatically} (see column 4). This suggests that many developers of social media datasets tend to neglect the data deduplication process. In addition to duplicated samples, we also observe a substantial number of near-duplicate samples in each dataset (see the last column).

\section{Impact of Duplicate and Near-duplicate Samples}
\label{experiments}

We conduct comparative experiments regarding three potential issues (data leakage, model rankings and inconsistent labels) on the selected social media datasets to verify the impact of those duplicate and near-duplicate posts on classification performance.

\begin{table*}[!t]
\small
\centering
\begin{tabular}{|l|c|c|c|c|c|c|}
\hline
\rowcolor{dGray} & \multicolumn{2}{|c|}{\textbf{Original}} & \multicolumn{2}{|c|}{\textbf{w/o Duplicates}} & \multicolumn{2}{|c|}{\textbf{w/o Near-Duplicates}}\\
\rowcolor{dGray} \textbf{Dataset} & \textbf{F1} & \textbf{Acc.} & \textbf{F1} & \textbf{Acc.} & \textbf{F1} & \textbf{Acc.}\\
\hline
\rowcolor{mGray} \multicolumn{7}{|c|}{\textbf{Offensive Language Detection}}\\
\hline
\textbf{\texttt{WASEEM}} & 84.2$\pm$0.4 & 86.2$\pm$0.3 & 83.4$\pm$0.4 & 85.7$\pm$0.3 & 83.0$\pm$0.2 & 85.3$\pm$0.3\\
\hline
\textbf{\texttt{TBO}} & 69.4$\pm$1.7 & 75.0$\pm$0.7 & 69.4$\pm$1.7 & 75.0$\pm$0.7 & 69.2$\pm$1.1 & 74.7$\pm$0.3\\
\hline
\textbf{\texttt{OLID}} & 76.4$\pm$0.5 & 79.9$\pm$0.1 & 77.1$\pm$0.4 & 80.1$\pm$0.4& 76.8$\pm$0.6& 79.0$\pm$0.9\\
\hline
\textbf{\texttt{FOUNTA}} & 85.9$\pm$0.0 & 86.0$\pm$0.0 &85.7$\pm$0.2 & 85.9$\pm$0.1 & 85.9$\pm$0.1 & 86.0$\pm$0.1\\
\hline
\textbf{\texttt{HateEval'19}} & 78.2$\pm$1.0 & 78.7$\pm$0.6 & 80.4$\pm$0.3 & 80.8$\pm$0.1 & 80.2$\pm$0.3& 80.8$\pm$0.2\\
\hline
\rowcolor{mGray} \multicolumn{7}{|c|}{\textbf{Misinformation Detection}}\\
\hline
\textbf{\texttt{Twitter'15}} & 60.4$\pm$2.2 & 61.0$\pm$1.7 & 58.4$\pm$0.5\textbf{*} & 58.7$\pm$0.3\textbf{*} & 57.1$\pm$3.2\textbf{*}&57.7$\pm$2.4\textbf{*}\\
\hline
\textbf{\texttt{Twitter'16}} & 62.6$\pm$3.5 & 63.4$\pm$3.1 & 51.9$\pm$4.0\textbf{*} & 55.7$\pm$2.5\textbf{*} & 43.0$\pm$2.2\textbf{*} &50.2$\pm$1.9\textbf{*}\\
\hline
\textbf{\texttt{PHEME}} & 84.7$\pm$0.7 & 86.2$\pm$0.4 & 84.1$\pm$0.4\textbf{*} & 85.6$\pm$0.4\textbf{*} &83.8$\pm$0.2\textbf{*}&85.5$\pm$0.0\textbf{*}\\
\hline
\textbf{\texttt{Weibo}} & 91.4$\pm$0.5 & 91.4$\pm$0.4 &90.9$\pm$0.4\textbf{*} & 90.9$\pm$0.4\textbf{*} & 91.0$\pm$0.2\textbf{*}&91.0$\pm$0.2\textbf{*}\\
\hline
\textbf{\texttt{STANKER}} & 92.2$\pm$0.2 & 92.2$\pm$0.2 & 91.7$\pm$0.2 & 91.7$\pm$0.2 &92.6$\pm$0.2&92.6$\pm$0.2\\
\hline
\rowcolor{mGray} \multicolumn{7}{|c|}{\textbf{Speech Act \& Sentiment Analysis}}\\
\hline
\textbf{\texttt{Complaint}} & 88.9$\pm$1.3 & 89.8$\pm$1.3 & 89.5$\pm$1.3 & 90.3$\pm$1.2 & 89.2$\pm$0.3&90.0$\pm$0.2\\
\hline
\textbf{\texttt{Sarcasm}} & 64.7$\pm$7.6 & 81.5$\pm$1.0 & 61.4$\pm$12.4 & 80.8$\pm$1.9 & 69.8$\pm$0.8&82.2$\pm$0.7\\
\hline
\textbf{\texttt{Bragging}} & 77.8$\pm$0.7 & 91.4$\pm$0.5 & 77.2$\pm$0.5 & 91.1$\pm$0.3 & 77.8$\pm$0.7&91.3$\pm$0.7\\
\hline
\textbf{\texttt{Parody}} & - & - & - & - & - & -\\
\hline
\textbf{\texttt{Sentiment}} & 74.7$\pm$0.2 & 75.0$\pm$0.3 & 74.5$\pm$0.2 & 75.0$\pm$0.4&74.5$\pm$0.3&74.9$\pm$0.3\\
\hline
\rowcolor{mGray} \multicolumn{7}{|c|}{\textbf{Stance Detection}}\\
\hline
\textbf{\texttt{CovidVaxx}} & 80.7$\pm$0.4 & 80.7$\pm$0.5 & 79.2$\pm$0.9\textbf{*} & 79.2$\pm$0.9\textbf{*} & 79.0$\pm$0.8\textbf{*}&79.1$\pm$0.8\textbf{*}\\
\hline
\textbf{\texttt{RumorEval}} & 48.6$\pm$1.2 & 72.7$\pm$0.5 & 48.6$\pm$0.4 & 72.8$\pm$0.5&46.4$\pm$0.9&72.4$\pm$1.1\\
\hline
\textbf{\texttt{US-Election}} & 53.4$\pm$0.7 & 56.8$\pm$0.9 & 52.4$\pm$0.8 & 56.5$\pm$0.8 &51.2$\pm$1.7&55.3$\pm$0.5\\
\hline
\textbf{\texttt{P-Stance}} & 80.1$\pm$0.7 & 80.2$\pm$0.7 & 80.8$\pm$0.4 & 80.9$\pm$0.4 &80.8$\pm$0.2&80.9$\pm$0.2\\
\hline
\textbf{\texttt{SemEval'16}} & 65.9$\pm$0.3 & 69.5$\pm$0.5 & 65.7$\pm$0.6 & 69.1$\pm$0.4 & 63.1$\pm$1.1\textbf{*}&67.4$\pm$1.2\textbf{*}\\
\hline
\end{tabular}
\caption{Model performance across datasets and duplicate rates. We were unable to conduct experiments on the Parody dataset due to the incomplete dataset we obtained. \textbf{*} denotes the statistic significance ($t$-test, $p$ < .05) between original and w/o settings. We run all BERT-style models three times with different random seeds and then report the average F1 measure and accuracy.}
\label{tab:results}
\end{table*}

\begin{figure*}[t!]
     \centering
     \begin{subfigure}[b]{0.49\textwidth}
         \centering
         \includegraphics[width=\textwidth]{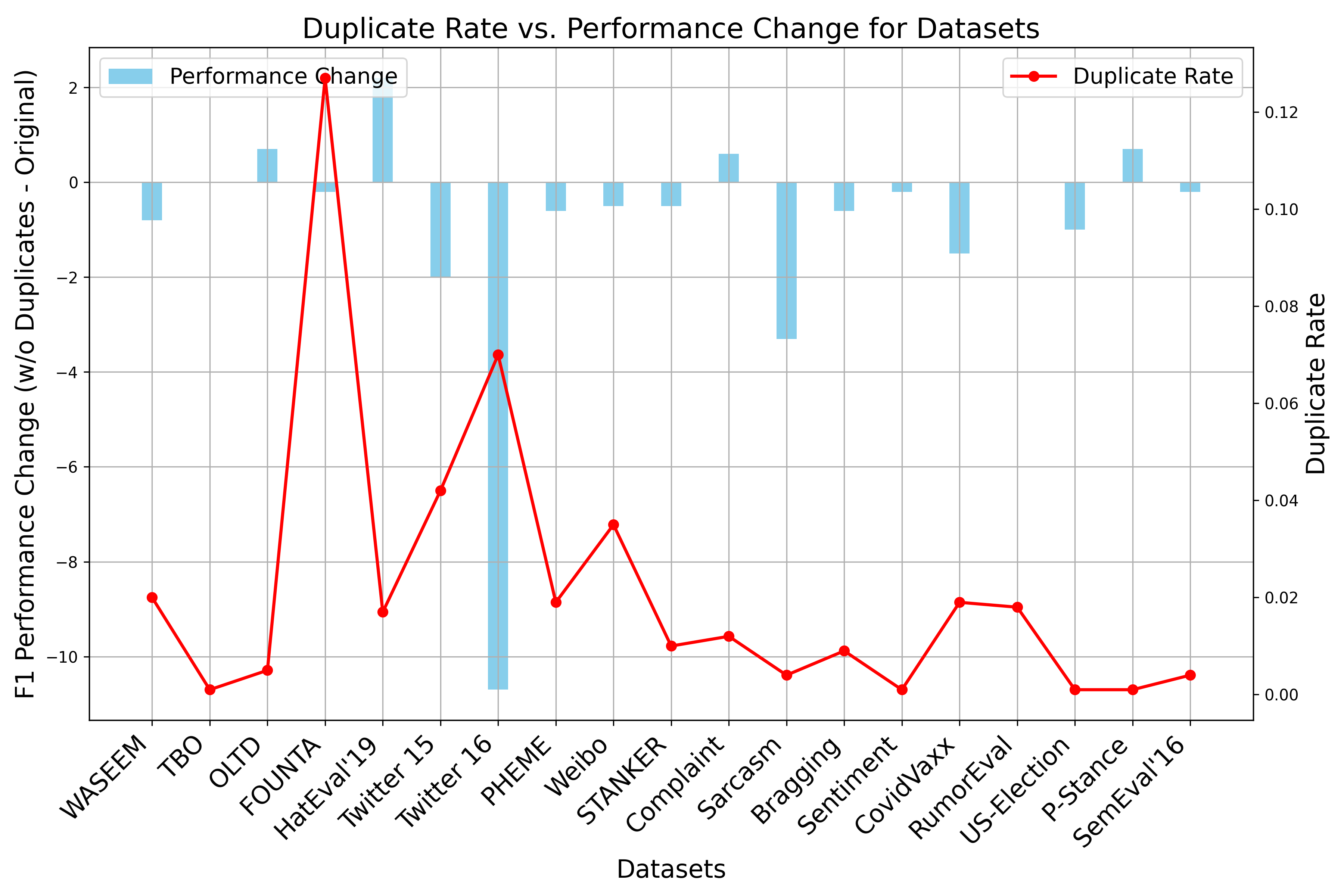}
         \label{fig:waseem_topic}
     \end{subfigure}
     \hfill
     \begin{subfigure}[b]{0.49\textwidth}
         \centering
         \includegraphics[width=\textwidth]{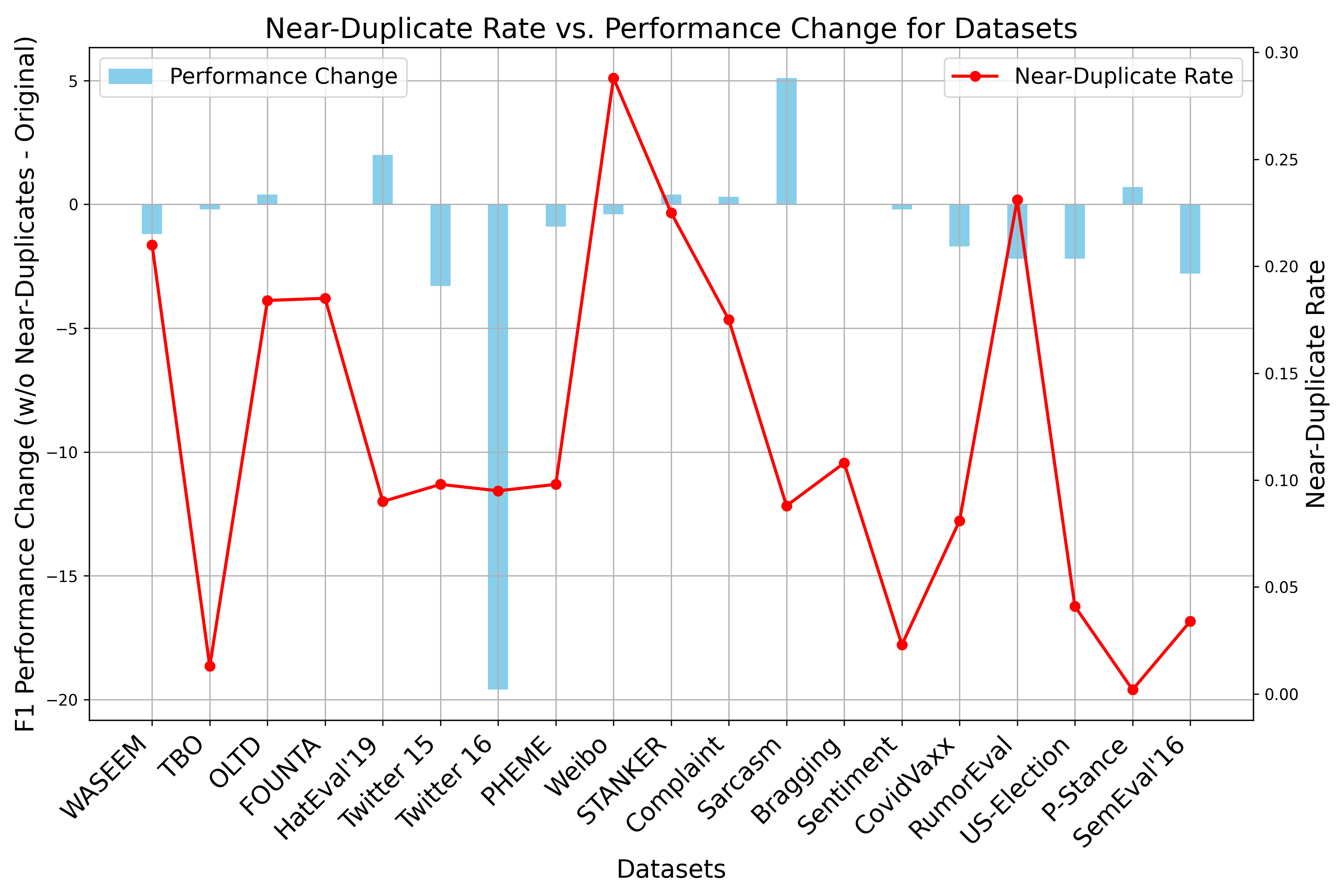}
         \label{fig:founta_topic}
     \end{subfigure}
        \caption{Duplicate (left) and Near-Duplicate (right) rates against performance changes in F1 scores. The positive values on the left y-axis (performance change) indicate an improvement while the negative values on the left y-axis indicate a decline. The zero line marks no change in performance.}
        \label{fig:performance_change}
\end{figure*}

\subsection{Data Leakage}
Data leakage occurs because the model is evaluated on data it has already seen during training. When test data overlaps with training data (e.g., the dataset contains duplicates), the performance of the model is likely to be overestimated. To investigate the impact of data leakage on model performance, we compare the predictive results within a dataset before and after deduplication.

We use BERTweet \citep{nguyen2020bertweet} for all English datasets and Bert-base-Chinese \citep{devlin2019bert} for two Chinese datasets. For all datasets, we first replace user mentions and URLs with special tokens. Then we split each dataset into a training (80\%) (\textit{Train Set Original}) and a test set (20\%). We construct \textit{Train Set w/o Duplicates} by removing all posts in the training set that are identical to those in the test set. Similarly, we build \textit{Train Set w/o Near-duplicate} using the same approach, removing near-duplicate posts. This deduplication process aims to prevent label leakage during model training. It is important to note that the test set remains unchanged for fair comparison. Detailed experimental setup is presented in Appendix \ref{appendix:parameter}.

With the widespread use of large language models (LLMs), we aim to compare their predictive results with traditional pre-trained models that require training data (training on duplicates potentially leads to data leakage). To facilitate this, we employ the two most recent LLMs, GPT-4o\footnote{\url{https://platform.openai.com/docs/models/gpt-4o}} and LLaMA 3-8B-Instruct\footnote{\url{https://huggingface.co/meta-llama/Meta-Llama-3-8B-Instruct}} for zero-shot classification. For LLMs, we use the same test set as for BERT-style models. Note that we only evaluate LLMs on the task of \texttt{Complaint}, \texttt{Bragging}, \texttt{Sarcasm}, \texttt{CovidVaxx} and \texttt{US-Election} since the sizes of other datasets are too large (see Table \ref{tab:datasets}). Example prompts are provided in Appendix \ref{appendix: prompt_llm}.

\subsection{Model Rankings}

The prevalence of duplicates and near-duplicates in these datasets prompts us to investigate the consistency of model rankings before and after deduplication. Given the extensive workload of reproducing all baseline and state-of-the-art approaches, we propose a new evaluation method. Firstly, we save 5 model checkpoints from the last 5 epochs during training, each representing a unique model with varying parameters. We then compare the ranking of these 5 models according to F1-macro scores before and after deduplication. We use \textit{Train Set Original} and \textit{Train Set w/o Duplicates} for training and the same test set for testing, similar to the previous experiment.

\subsection{Inconsistent Labels}

In the few-shot experiment, we aim to investigate the impact of label inconsistency in duplicates on the predictive performance of LLMs using two similar few-shot prompts:
\begin{itemize}
    \item \textit{Given Sample A (label X) and Sample A (label Y), predict if the following example is X or Y: Sample A};

    \item \textit{Given Sample A (label Y) and Sample A (label X), predict if the following example is X or Y: Sample A}.
\end{itemize}

 We then insert $n$ randomly selected samples between two Sample A instances with different labels for both prompts, where $n$ varies from 0 to 5. For comparison, we also use zero-shot evaluation with prompt \textit{predict if the following example is X or Y: Sample A}.

\begin{table*}[!t]
\small
\centering
\begin{tabular}{|l|ll|ll|ll||ll|ll|}
\hline
\rowcolor[HTML]{C0C0C0} 
 &
  \multicolumn{2}{c|}{\cellcolor[HTML]{C0C0C0}\textbf{Original}} &
  \multicolumn{2}{c|}{\cellcolor[HTML]{C0C0C0}\textbf{w/o Dupl.}} &
  \multicolumn{2}{c||}{\cellcolor[HTML]{C0C0C0}\textbf{w/o Near-dupl.}} &
  \multicolumn{2}{c|}{\cellcolor[HTML]{C0C0C0}\textbf{GPT-4o}} &
  \multicolumn{2}{c|}{\cellcolor[HTML]{C0C0C0}\textbf{LLaMA-3}} \\ \hline
\rowcolor[HTML]{C0C0C0} 
\textbf{Dataset} &
  \multicolumn{1}{c|}{\cellcolor[HTML]{C0C0C0}\textbf{F1}} &
  \multicolumn{1}{c|}{\cellcolor[HTML]{C0C0C0}\textbf{Acc.}} &
  \multicolumn{1}{c|}{\cellcolor[HTML]{C0C0C0}\textbf{F1}} &
  \multicolumn{1}{c|}{\cellcolor[HTML]{C0C0C0}\textbf{Acc.}} &
  \multicolumn{1}{c|}{\cellcolor[HTML]{C0C0C0}\textbf{F1}} &
  \multicolumn{1}{c||}{\cellcolor[HTML]{C0C0C0}\textbf{Acc.}} &
  \multicolumn{1}{c|}{\cellcolor[HTML]{C0C0C0}\textbf{F1}} &
  \multicolumn{1}{c|}{\cellcolor[HTML]{C0C0C0}\textbf{Acc.}} &
  \multicolumn{1}{c|}{\cellcolor[HTML]{C0C0C0}\textbf{F1}} &
  \multicolumn{1}{c|}{\cellcolor[HTML]{C0C0C0}\textbf{Acc.}} \\ \hline
\textbf{\texttt{Complaint}} &
  \multicolumn{1}{l|}{88.9±1.3} &
  89.8±1.3 &
  \multicolumn{1}{l|}{89.5±1.3} &
  90.3±1.2 &
  \multicolumn{1}{l|}{89.2±0.3} &
  90.0±0.2 &
  \multicolumn{1}{l|}{81.2} &
  81.4 &
  \multicolumn{1}{l|}{77.4} &
  77.7 \\ \hline
\textbf{\texttt{Bragging}} &
  \multicolumn{1}{l|}{77.8±0.7} &
  91.4±0.5 &
  \multicolumn{1}{l|}{77.2±0.5} &
  91.1±0.3 &
  \multicolumn{1}{l|}{77.8±0.7} &
  91.3±0.7 &
  \multicolumn{1}{l|}{70.0} &
  85.2 &
  \multicolumn{1}{l|}{47.9} &
  62.5 \\ \hline
\textbf{\texttt{Sarcasm}} &
  \multicolumn{1}{l|}{64.7±7.6} &
  81.5±1.0 &
  \multicolumn{1}{l|}{61.4±12.4} &
  80.8±1.9 &
  \multicolumn{1}{l|}{69.8±0.8} &
  82.2±0.7 &
  \multicolumn{1}{l|}{53.1} &
  53.9 &
  \multicolumn{1}{l|}{39.4} &
  59.1 \\ \hline
\textbf{\texttt{CovidVaxx}} &
  \multicolumn{1}{l|}{80.7±0.4} &
  80.7±0.5 &
  \multicolumn{1}{l|}{79.2±0.9} &
  79.2±0.9 &
  \multicolumn{1}{l|}{79.0±0.8} &
  79.1±0.8 &
  \multicolumn{1}{l|}{77.6} &
  76.9 &
  \multicolumn{1}{l|}{63.6} &
  63.1 \\ \hline
\textbf{\texttt{US-Election}} &
  \multicolumn{1}{l|}{53.4±0.7} &
  56.8±0.9 &
  \multicolumn{1}{l|}{52.4±0.8} &
  56.5±0.8 &
  \multicolumn{1}{l|}{51.2±1.7} &
  55.3±0.5 &
  \multicolumn{1}{l|}{47.7} &
  50.0 &
  \multicolumn{1}{l|}{41.5} &
  46.4 \\ \hline
\end{tabular}%
\caption{Predictive results of LLMS (GPT-4o and LLaMA-3-8B) (right) vs. BERTweet (Original, w/o Dupl. and w/o Near-dupl.) (left).}
\label{tab:llm_results}
\end{table*}

\begin{table*}[!t]
\small
\centering
\resizebox{0.97\textwidth}{!}{

\begin{tabular}{|l|c|c|c|c|c|}

\hline
\rowcolor{dGray}\textbf{Dataset} & \textbf{NO.1} & \textbf{NO.2} & \textbf{NO.3} & \textbf{NO.4} & \textbf{NO.5} \\  
\hline
\texttt{\textbf{HatEval'19}}-duplicate & 80.25 (E6) & 80.11 (E7) & 80.01 (E10) & 79.86 (E7) & 79.41 (E9)\\
\texttt{\textbf{HatEval'19}}-noduplicate & 80.18 (E9) & 80.01 (E10) & 79.78 (E8) & 79.77 (E7) &79.34 (E6)\\
\hline
\texttt{\textbf{Twitter 16}}-duplicate & \cellcolor{lGray} 75.30 (E10) & \cellcolor{lGray} 74.65 (E9) & \cellcolor{lGray} 74.56 (E8) & \cellcolor{lGray} 73.97 (E7) & \cellcolor{lGray} 73.44 (E6)\\
\texttt{\textbf{Twitter 16}}-noduplicate & \cellcolor{lGray} 67.22 (E10) & \cellcolor{lGray} 67.04 (E9) & \cellcolor{lGray} 66.92 (E8) & \cellcolor{lGray} 66.40 (E7) & \cellcolor{lGray} 66.22 (E6)\\
\hline
\texttt{\textbf{Bragging}}-duplicate & 80.00 (E7) & 79.45 (E9) & 79.33 (E8) &78.60 (E10) &78.47 (E6)\\
\texttt{\textbf{Bragging}}-noduplicate & 78.22 (E6) & 78.00 (E7) & 77.86 (E9) & 77.77 (E8) & 77.39 (E10)\\
\hline 
\texttt{\textbf{P-Stance}}-duplicate & 81.17 (E10) & 80.80 (E9) & 80.72 (E8) & 80.71 (E6) & 80.39 (E7) \\
\texttt{\textbf{P-Stance}}-noduplicate & 80.55 (E8) & 80.38 (E10) & 80.36 (E7) & 80.35 (E9) &80.14 (E6) \\
\hline 
\end{tabular}
}
\caption{Rankings of five model checkpoints across selected datasets based on macro-F1 scores. E denotes epoch, e.g., E6 refers to 6th Epoch. Unchanged rankings are in light gray. Full experimental results are displayed in Appendix \ref{appendix:ranking}.}
\label{tab:ranking}
\end{table*}

\subsection{Results and Discussion}
\paragraph{Data Leakage.}
Table \ref{tab:results} presents the predictive results across all tasks with different training set configurations (i.e., Original, w/o Duplicates and w/o Near-Duplicates) and Figure \ref{fig:performance_change} shows the duplicate (left) and near-duplicate (right) rates against F1 score changes. In general, we observe that the presence of duplicated samples in the training set results in an overestimation of the model's predictive performance (14 out of 19 datasets). For example, the model achieves 62.6 F1 in `Original' and 51.9 F1 in `w/o Duplicates' ($t$-test, $p$ < .05) in \texttt{Twitter 16}. This is due to duplicate posts in the training and test set leading to label leakage during model training.
Meanwhile, we notice that model performance on datasets (e.g., \texttt{Sentiment} and \texttt{P-Stance}) with minimal duplicates (i.e., Duplicated\% < 0.01) remains steady. Additionally, we observe a similar result in `w/o Duplicates' and `w/o Near-duplicate' where the results of the latter are slightly worse. 

Table \ref{tab:llm_results} demonstrates that the zero-shot classification performance of recent LLMs may not always surpass that of fully fine-tuned BERT-style models. This suggests that supervised approaches remain indispensable in CSS research, thereby making the resolution of the data duplication issue inevitable.

\paragraph{Model Rankings.}
We present cross-model evaluation results based on macro-F1 scores for selected datasets in Table \ref{tab:ranking} and for all datasets in Appendix \ref{appendix:ranking}. The findings reveal that the majority of model rankings exhibit inconsistency (17 out of 19 datasets) before and after deduplication. Taking \texttt{HatEval'19} for example, the top five model checkpoints are from Epoch 6, 7, 10, 7 and 9 respectively when containing duplicates; while the top five ones are from Epoch 9, 10, 8, 7 and 6 after deduplication. However, we notice the same rankings for dataset \texttt{Twitter 16} and \texttt{TBO}, which may result from the small size of the dataset (e.g., 818 tweets). This suggests that data duplication undermines the validity of claimed SoTA results. We contend that current SoTA approaches may require reassessment through essential data cleaning measures.

\paragraph{Inconsistent Labels.}

Table \ref{tab:fewshot} presents the results of stance detection using few-shot prompts. We observe that the model can predict the stance correctly in zero-shot settings. Also, when presented with fewer instances containing posts with inconsistent labels, the model tends to make more accurate predictions. However, when provided with more than five instances, the model begins to align its predictions with the most recent label. This indicates that as the number of instances increases, predictions of the model may be less reliable, particularly when presented with conflicting information. We obtain similar findings from other social media tasks. This highlights the importance of data quality and consistency in training models for downstream CSS tasks.

\begin{table}[!t]
\resizebox{0.49\textwidth}{!}{
\small
\begin{tabular}{|l|p{1.1cm}|}

\hline
\rowcolor{dGray} \textbf{Prompt} & \textbf{Pred.} \\   
\hline
Zero-shot & Against \\  
\hline
Sample A (Neutral)  & Against\\               
Sample A (Against) & Against \\  
\hline  
Sample A (Neutral), Sample A (Against)  & Against\\
Sample A (Against), Sample A (Neutral) & Neutral \\  
\hline    
Sample A (Neutral), Sample B, Sample A (Against)  & Against\\
Sample A (Against), Sample B, Sample A (Neutral) & Against \\  
\hline 
Sample A (Neutral), Sample B, C, Sample A (Against)  & Against\\
Sample A (Against), Sample B, C, Sample A (Neutral) & Against \\  
\hline 
Sample A (Neutral), Sample B, C, D, Sample A (Against)  & Against\\
Sample A (Against), Sample B, C, D, Sample A (Neutral) & Neutral \\  
\hline 
Sample A (Neutral), Sample B, C, D, E, Sample A (Against)  & Against\\
Sample A (Against), Sample B, C, D, E, Sample A (Neutral) & Neutral \\  
\hline 
Sample A (Neutral), Sample B, C, D, E, F, Sample A (Against)  & Against\\
Sample A (Against), Sample B, C, D, E, F, Sample A (Neutral) & Neutral \\  
\hline 
\end{tabular}
}
\caption{Predictive results of GPT-4o on stance detection using US-Election. Sample A is \textit{@USER @USER donald trump's lessons for republicans: no consequences for lying. no consequences for hate. no consequences for immorality. no consequences for crimes. no consequences for collusion. no consequences for impeachment. no consequences for incompetence. sad. \#votebluenomatterwho} and Sample B, C, D, E, F are randomly selected with original labels from the same dataset.}
\label{tab:fewshot}
\end{table}

\subsection{Error Analysis}
To further investigate the impact of such duplicates and near-duplicates in social media datasets, we manually perform an error analysis on the test sets of five misinformation detection datasets. We first mark the duplicates and near-duplicates in the test set based on \textit{Train Set Original}. Figure \ref{fig:wrong_predictions} presents the percentage of these duplicates (upper) and these near-duplicates (bottom) in wrong predictions by models from five misinformation detection datasets (see Appendix \ref{appendix:error}). We observe that most duplicated ($\sim$ 99\%) and near-duplicated (more than 90\%) samples across the training and test sets can be correctly predicted.

This emphasizes the potential overestimation of model performance on existing social media datasets as a result of label leakage. Nonetheless, the presence of duplicate posts can still pose challenges for the model, especially when there is label inconsistency, as demonstrated in Table \ref{tab:duplicates_examples}. Our error analysis highlights the adverse impact of duplicated and near-duplicated samples on the reliability of model performance.

\section{Effective Strategies for Development and Use of Social Media Datasets}
\label{suggestions}

This section outlines practical recommendations we propose for more effective development and use of datasets in responsible CSS research.

\subsection{Developing New Datasets}

Developing a new social media dataset typically involves several key steps in a well-defined pipeline including problem definition, data collection, annotation and labeling, preprocessing, optionally balancing and stratification.

However, we suggest performing an initial data cleaning before the annotation task. More specifically, we recommend first performing standard data preprocessing steps (e.g., replacing @USER and URL tokens)\footnote{For annotation tasks, developers may want to restore these tokens for providing the original posts to annotators.} and then removing duplicate samples. This process can help reduce human annotation costs \citep{nguyen2017automatic} and avoid potentially inconsistent labeling. Additionally, manual rules (e.g., excluding posts with fewer than $N$ tokens) can also be used for further data cleaning before annotating, as suggested by \citet{zampieri2023target} who developed a dataset with almost no duplicates (see Table \ref{tab:datasets}).

Also, we suggest dataset developers provide different deduplication versions of datasets, for example, including and excluding near-duplicate posts, as both of them are likely to provide valuable information for the community.

\subsection{Using Existing Datasets}
An existing dataset may be used for model training as is, or enriched through fine-grained reannotation or incorporating multiple modalities. To make the most of existing datasets, attention to data quality is crucial, especially when dealing with duplicate samples. We suggest retaining one of the duplicate posts with consistent labels and excluding all posts with conflicting labels unless a reliable re-annotation task is performed. In cases where duplicates are absent, consider employing out-of-box similarity checks using methods like Levenshtein Distance and Cosine Similarity to exclude near-duplicate samples.

Notably, for datasets like PHEME annotated with more than one type of label (e.g., Veracity label: \textit{Rumor} or \textit{Non-rumor} and Event-related label: \textit{Charlie Hebdo Shooting} or \textit{Ottawa Shooting}, consider task-specific data splitting strategies. The leave-one-out protocol \citep{lukasik2016hawkes}, for instance, is advocated for more representative and effective evaluations.

\subsection{Updating Data Checklists}
\label{checklist}
Recently, most Computer Science venues have mandated a Data \& Ethics checklist to promote \textit{Responsible AI Research} emphasizing ethical and reproducibility considerations \citep{dodge2019show,rogers2021just}.

We argue that current versions of Data \& Ethics checklists from two leading organizations in CSS, *ACL\footnote{\url{https://aclrollingreview.org/responsibleNLPresearch/}} and AAAI-ICWSM\footnote{\url{https://www.overleaf.com/latex/templates/aaai-icwsm-2024-paper-checklist/vxbztbhhrbch}}, may need a `minor revision' to account for data deduplication. Therefore, we recommend including an additional question to remind researchers regarding the essential process of deduplicating data. Below are our revised versions for the two Data \& Ethics checklists:
\begin{quote}
      \textbf{Under *ACL Checklist Section B}
    
    \textbf{\textit{Title: Did you use or create scientific artifacts?}}

    Did you perform dataset exploration, such as applying data pre-processing and \textbf{deduplication}, when creating new or utilizing existing data resources?
\end{quote}

\begin{quote}
    \textbf{Under AAAI-ICWSM Checklist Section 5}
    
    \textbf{\textit{Title: If you are using existing assets (e.g., code, data, models) or curating/releasing new assets.}}
    
    Have you provided details of any meta-analysis conducted on the CSS datasets employed or developed in your research? This includes, but is not limited to, data pre-processing, \textbf{deduplication}, and other processes essential for responsible CSS research.
\end{quote}
We believe this adjustment applies not only to CSS but also to broader research fields.

\section{Conclusion}
Beyond focusing on innovative methodologies to improve model performance in downstream CSS tasks, we advocate a more comprehensive understanding of the datasets via meta analysis. In this work, we conduct an extensive evaluation of the data duplication issue in 20 selected social media datasets. We find that the majority of these datasets require additional preprocessing before the modeling, a critical step often overlooked in previous CSS research. Furthermore, we explore the potential impact of data duplication on model performance with a focus on label leakage, model rankings and label inconsistency. Finally, we offer targeted recommendations for more effective use of existing datasets and the creation of new datasets derived from social media sources.

\section*{Ethics Statement}
The Research Ethics Committee of our institute has granted approval for our work. The datasets evaluated in this study were acquired from original authors by request or via links available in the source papers. 

\section*{Limitations}
Due to space limits, our work only focuses on 20 representative CSS datasets, which do not cover the entire scope of computational social science. However, our work has examined a similar or higher number of datasets compared to existing CSS benchmarks such as TweetEval \citep{barbieri2020tweeteval} and SuperTweetEval \citep{antypas2023supertweeteval}. In the future, we are committed to continually expanding our analysis by evaluating more datasets, with updates to be shared through our GitHub repository. We aim to create a processed CSS benchmark, similar to existing CSS benchmarks such as SuperTweetEval \citep{antypas2023supertweeteval}. This will include a cleaned version of each dataset, along with re-evaluated baselines based on these deduplicated datasets.

\section*{Acknowledgments}
This work has been supported by the UK’s innovation agency (Innovate UK) grant 10039055 (approved under the Horizon Europe Programme as vera.ai, EU grant agreement 101070093). NA is supported by EPSRC grant EP/Y009800/1, part of the RAI UK Keystone projects.

\bibliography{custom}
\appendix
\section*{Appendix}
\section{Hyper-parameters and Experimental Setup}
\label{appendix:parameter}

Following the standard pipeline \cite{devlin2019bert}, we fine-tune BERT-style models by feeding the `[CLS]' token to a linear classifier with Softmax activation. We set the learning rate $lr$ = 2e-5 and batch size $bs$ = 64 for all datasets. All BERT variants are fine-tuned for up to 10 epochs using an early stopping strategy based on the validation loss (we use 10\% of the data from the training set as validation sets for model selection). We run all BERT-style models three times with different random seeds and then report the standard deviations and average Precision, Recall, and F1-measure. For LLMs, we fix the random seed and temperature values to ensure reproducibility.

All supervised experiments are conducted on an Nvidia V100 GPU with 32 GB of memory, with a total running time of approximately 3 hours (on 20 datasets described in Table \ref{tab:datasets}). All GPT prompting experiments cost around 3.5 USD in total, which can be fully covered by the free budget provided by OpenAI for new users.

\section{Complete results of model rankings}
\label{appendix:ranking}
Table \ref{tab:appendix_ranking} shows the full results of rankings of five model checkpoints across all datasets.

\section{Error Analysis}
\label{appendix:error}
\begin{figure}[!t]
  \centering
  \includegraphics[width=0.40\textwidth]{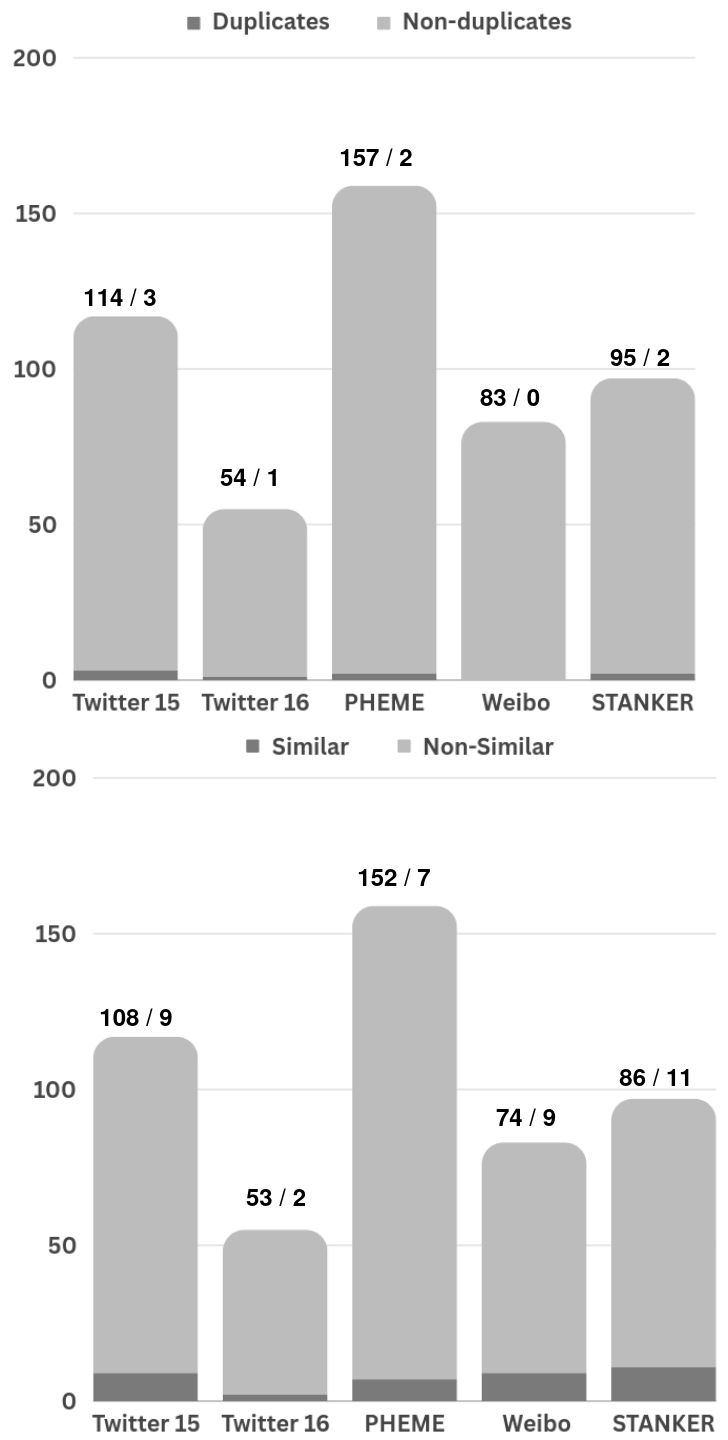}
  \caption{\small Ratio of duplicates (upper) and near-duplicates (bottom) in wrong predictions from five misinformation detection datasets.}
  \label{fig:wrong_predictions}
\end{figure}

Figure \ref{fig:wrong_predictions} presents the percentage of these duplicates (upper) and these near-duplicates (bottom) in wrong predictions by models from five misinformation detection datasets.

\begin{table*}[!t]
\centering
\resizebox{0.9\textwidth}{!}{

\begin{tabular}{|l|c|c|c|c|c|}

\hline
\rowcolor{dGray}\textbf{Dataset} & \textbf{NO.1} & \textbf{NO.2} & \textbf{NO.3} & \textbf{NO.4} & \textbf{NO.5} \\  
\hline
\multicolumn{6}{|c|}{\cellcolor{mGray} Offensive Language Detection} \\
\hline
\texttt{\textbf{WASEEM}}-duplicate & \cellcolor{lGray} 83.62 (E6) & 83.33 (E7) & 83.17 (E10) & \cellcolor{lGray} 83.06 (E9) & \cellcolor{lGray} 83.00 (E8)\\
\texttt{\textbf{WASEEM}}-noduplicate & \cellcolor{lGray} 83.71 (E6) & 82.89 (E10) & 82.61 (E7) & \cellcolor{lGray} 82.53 (E9) & \cellcolor{lGray} 82.51 (E8)\\
\hline
\texttt{\textbf{TBO}}-duplicate & \cellcolor{lGray} 69.76 (E6) & \cellcolor{lGray} 69.72 (E10) & \cellcolor{lGray} 69.63 (E9) & \cellcolor{lGray} 69.61 (E8) & \cellcolor{lGray} 69.54 (E7)\\
\texttt{\textbf{TBO}}-noduplicate& \cellcolor{lGray} 69.76 (E6) & \cellcolor{lGray} 69.72 (E10) & \cellcolor{lGray} 69.63 (E9) & \cellcolor{lGray} 69.61 (E8) & \cellcolor{lGray} 69.54 (E7)\\
\hline
\texttt{\textbf{OLID}}-duplicate & 75.30 (E7) & 74.66 (E6) & 74.52 (E9) & 74.38 (E10) & \cellcolor{lGray} 74.01 (E8)\\
\texttt{\textbf{OLID}}-noduplicate & 75.82 (E10) & 75.62 (E6) & 75.53 (E7) & 75.21 (E9) & \cellcolor{lGray} 75.05 (E8)\\
\hline
\texttt{\textbf{FOUNTA}}-duplicate & \cellcolor{lGray} 84.37 (E6) & 83.68 (E9) & 83.61 (E8) & 83.43 (E10) & 83.25 (E7)\\
\texttt{\textbf{FOUNTA}}-noduplicate & \cellcolor{lGray} 84.20 (E6) & 83.76 (E8) & 83.71 (E7) & 83.42 (E9) & 83.29 (E10)\\
\hline
\texttt{\textbf{HatEval'19}}-duplicate & 80.25 (E6) & 80.11 (E7) & 80.01 (E10) & \cellcolor{lGray} 79.86 (E7) & 79.41 (E9)\\
\texttt{\textbf{HatEval'19}}-noduplicate & 80.18 (E9) & 80.01 (E10) & 79.78 (E8) & \cellcolor{lGray} 79.77 (E7) &79.34 (E6)\\
\hline
\multicolumn{6}{|c|}{\cellcolor{mGray} Misinformation Detection} \\
\texttt{\textbf{Twitter 15}}-duplicate & 69.61 (E8) & 67.90 (E10) & 67.76 (E9) & 67.41 (E7) & 66.47 (E6) \\
\texttt{\textbf{Twitter 15}}-noduplicate & 67.58 (E7) & 67.19 (E8) & 67.00 (E10) & 66.75 (E9) & 65.45 (E6)\\
\hline
\texttt{\textbf{Twitter 16}}-duplicate & \cellcolor{lGray} 75.30 (E10) & \cellcolor{lGray} 74.65 (E9) & \cellcolor{lGray} 74.56 (E8) & \cellcolor{lGray} 73.97 (E7) & \cellcolor{lGray} 73.44 (E6)\\
\texttt{\textbf{Twitter 16}}-noduplicate & \cellcolor{lGray} 67.22 (E10) & \cellcolor{lGray} 67.04 (E9) & \cellcolor{lGray} 66.92 (E8) & \cellcolor{lGray} 66.40 (E7) & \cellcolor{lGray} 66.22 (E6)\\
\hline
\texttt{\textbf{PHEME}}-duplicate & \cellcolor{lGray} 86.93 (E8) & 86.58 (E10) & 86.54 (E7) & 86.54 (E9) & \cellcolor{lGray} 85.92 (E6)\\
\texttt{\textbf{PHEME}}-noduplicate & \cellcolor{lGray} 85.95 (E8) & 85.91 (E10) & 85.83 (E9) & 85.80 (E7) & \cellcolor{lGray} 85.61 (E6)\\
\hline
\texttt{\textbf{Weibo}}-duplicate & 91.75 (E9) & 91.43 (E10) & 91.10 (E8) & 90.78 (E7) & 90.35 (E6)\\
\texttt{\textbf{Weibo}}-noduplicate & 91.75 (E8) & 91.64 (E9) & 91.00 (E10) & 90.68 (E6) & 90.25 (E7)\\
\hline
\texttt{\textbf{STANKER}}-duplicate & 93.82 (E7) & 93.66 (E8) & 93.33 (E9) & 93.24 (E10) & 93.16 (E6)\\
\texttt{\textbf{STANKER}}-noduplicate & 92.92 (E10) & 92.83 (E9) & 92.75 (E8) & 92.25 (E6) & 92.17 (E7)\\
\hline
\multicolumn{6}{|c|}{\cellcolor{mGray} Speech Act \& Sentiment Analysis} \\
\texttt{\textbf{Complaint}}-duplicate & 90.73 (E6) & 90.40 (E9) & 90.34 (E10) & 90.01 (E8) & 89.60 (E7)\\
\texttt{\textbf{Complaint}}-noduplicate & 90.85 (E7) & 90.59 (E10) & 90.52 (E6, E9) & 90.52 (E6, E9) & 90.44 (E8)\\
\hline
\texttt{\textbf{Sarcasm}}-duplicate & 71.53 (E10) & 71.45 (E9) & 71.21 (E8) & 70.76 (E6) & 70.22 (E7)\\
\texttt{\textbf{Sarcasm}}-noduplicate & 72.41 (E7) & 71.79 (E10) & 71.64 (E6) & 71.49 (E8) & 71.42 (E9)\\
\hline
\texttt{\textbf{Bragging}}-duplicate & 80.00 (E7) & 79.45 (E9) & 79.33 (E8) &78.60 (E10) &78.47 (E6)\\
\texttt{\textbf{Bragging}}-noduplicate & 78.22 (E6) & 78.00 (E7) & 77.86 (E9) & 77.77 (E8) & 77.39 (E10)\\
\hline 
\texttt{\textbf{Parody}}-duplicate & - & - & - & - &\\
\texttt{\textbf{Parody}}-noduplicate & - & - & - & - &\\
\hline
\texttt{\textbf{Sentiment}}-duplicate & \cellcolor{lGray} 73.61 (E6) & 73.11 (E7) & 72.90 (E9) & 72.86 (E10) & 72.79 (E8)\\
\texttt{\textbf{Sentiment}}-noduplicate & \cellcolor{lGray} 73.39 (E6) & 72.90 (E9) & 72.81 (E10) & 72.78 (E8) & 72.56 (E7)\\
\hline
\multicolumn{6}{|c|}{\cellcolor{mGray} Stance Detection} \\
\hline
\texttt{\textbf{CovidVaxx}}-duplicate & 80.93 (E8) & 80.78 (E6) & 80.43 (E9) & 79.89 (E10) & 78.99 (E7)\\
\texttt{\textbf{CovidVaxx}}-noduplicate & 81.01 (E7) & 80.32 (E9) & 80.27 (E6) & 80.27 (E8) & 80.11 (E10)\\
\hline
\texttt{\textbf{RumorEval}}-duplicate & 54.66 (E10) & 54.44 (E9) & 53.77 (E8) & 53.76 (E6) & \cellcolor{lGray} 53.42 (E7)\\
\texttt{\textbf{RumorEval}}-noduplicate & 54.27 (E6) & 54.03 (E10) & 53.96 (E9) & 53.83 (E8) & \cellcolor{lGray} 53.23 (E7)\\
\hline
\texttt{\textbf{US-Election}}-duplicate & \cellcolor{lGray} 56.53 (E8) & 56.04 (E7) & 54.90 (E9) & 54.63 (E6) & 54.12 (E10)\\
\texttt{\textbf{US-Election}}-noduplicate & \cellcolor{lGray} 56.20 (E8) & 55.36 (E6) & 55.56 (E10) & 55.46 (E9) & 52.85 (E7)\\
\hline
\texttt{\textbf{P-Stance}}-duplicate & 81.17 (E10) & 80.80 (E9) & 80.72 (E8) & 80.71 (E6) & 80.39 (E7) \\
\texttt{\textbf{P-Stance}}-noduplicate & 80.55 (E8) & 80.38 (E10) & 80.36 (E7) & 80.35 (E9) &80.14 (E6) \\
\hline 
\end{tabular}
}
\caption{Rankings of five model checkpoints across selected datasets based on macro-F1 scores. E denotes epoch, e.g., E6 refers to 6th Epoch. Unchanged rankings are in light gray.}
\label{tab:appendix_ranking}
\end{table*}

\section{Prompts for LLM Zero- and Few-shot Classification}
\label{appendix: prompt_llm}
\begin{table*}[]
\centering
\begin{tabular}{|l|l|}
\hline
\rowcolor[HTML]{C0C0C0} 
\multicolumn{1}{|c|}{\cellcolor[HTML]{C0C0C0}\textbf{Dataset}} &
  \multicolumn{1}{c|}{\cellcolor[HTML]{C0C0C0}\textbf{Prompt}} \\ \hline
Complaint &
  \begin{tabular}[c]{@{}l@{}}Read the given tweet, and categorize it into one of two categories: \\ (1) Non-complaint \\ (2) Complaint\\ \\ Only return the category number as your answer.\\ \\ Text: \{list\_of\_text\}\\ Answer:\end{tabular} \\ \hline
Bragging &
  \begin{tabular}[c]{@{}l@{}}Read the given tweet, and categorize it into one of two categories:\\ (1) Not Bragging\\ (2) Bragging\\ \\ Only return the category number as your answer.\\ \\ Text: \{list\_of\_text\}\\ Answer:\end{tabular} \\ \hline
Sarcasm &
  \begin{tabular}[c]{@{}l@{}}Read the given tweet, and categorize it into one of two categories:\\ (1) Not Sarcasm\\ (2) Sarcasm\\ \\ Only return the category number as your answer.\\ \\ Text: \{list\_of\_text\}\\ Answer:\end{tabular} \\ \hline
CovidVAXX &
  \begin{tabular}[c]{@{}l@{}}Read the given tweet, and categorize it according to the stance \\ expressed about the COVID-19 vaccine:\\ (1) Anti vaccine \\ (2) Neutral \\ (3) Pro vaccine \\ \\ Only return the category number as your answer.\\ \\ Text: \{list\_of\_text\}\\ Answer:\end{tabular} \\ \hline
US\_election &
  \begin{tabular}[c]{@{}l@{}}Read the given tweet, and categorize it according to the stance \\ expressed towards U.S. politicians:\\ (1) None\\ (2) Favor\\ (3) Against\\ \\ Only return the category number as your answer.\\ \\ Text: \{list\_of\_text\}\\ Answer:\end{tabular} \\ \hline
\end{tabular}
\caption{Example prompts used for LLM-based zero-shot classification.}
\label{tab:llm_promot}
\end{table*}

\end{document}